\documentclass[fleqn,a4]{article}
\usepackage{amsmath,amssymb,amscd}
\usepackage[pdftex]{graphicx}
\makeatletter

\selectfont

\renewcommand{\section}{\@startsection{section}{1}{\z@}%
{2ex}{1ex}{\reset@font\large\bfseries}}%
\renewcommand{\thesection}{\@arabic\c@section}
\def\@listi{\topsep=.3\baselineskip \parsep=.2ex \partopsep=0ex%
\itemsep=0ex \leftmargin=4ex \rightmargin=2ex}
\let\@listI\@listi
\@listi\def\@listii{\parsep=.2ex \partopsep=0pt \itemsep=0ex%
\leftmargin=4ex \rightmargin=0ex}
\let\@listiii\@listii
\let\@listiv\@listii
\let\@listv\@listii
\let\@listvi\@listii
\long\def\@makecaption#1#2{\footnotesize\sbox\@tempboxa{#1. #2}
\ifdim\wd\@tempboxa >\hsize #1. #2\par
\else \global\@minipagefalse
\hb@xt@\hsize{\hfil\box\@tempboxa\hfil}
\fi}
\makeatother
\usepackage{graphicx}
\usepackage{amsmath}
\usepackage{url}

\begin{document}
\title{The covariance matrix of Green's functions and its application to machine learning}

\author{Tomoko Nagai\\
Kogakuin University, 2665-1 Nakano-cho, Hachioji, Tokyo 192-0015, Japan.
} 


\maketitle
\begin{abstract}
In this paper, a regression algorithm based on Green's function theory is proposed and implemented.
We first survey Green's function for the Dirichlet boundary value problem of 2nd order linear ordinary 
differential equation, which is a reproducing kernel of a suitable Hilbert space. 
We next consider a covariance matrix composed of the normalized Green's function, which is regarded as a
probability density function. 
By supporting Bayesian approach, the covariance matrix gives
predictive distribution, which has the predictive mean $\mu$ and the confidence interval 
$[\mu-2s, \mu+2s]$, 
where $s$ stands for a standard deviation.
\end{abstract}
\section{Introduction}

Recently, it is shown that there is a close relationship between machine learning and reproducing kernel theory.\cite{fuku} 
A covariance matrix in Bayesian regression, which is composed of kernel functions, 
is presented as kernel matrix in regression. \cite{fuku,bishop,gp,kana} 
The Gaussian process regression and interpolation based on the Bayesian approach 
give the predictive mean and variance. \cite{bishop,gp,kana} 
On the other hand, it is proved that Green's functions to boundary value problem for differential equations, 
which are response functions for impulses, are reproducing kernels of suitable Hilbert spaces. \cite{kame, kame2} 
This fact suggests the relationship between Green's functions and kernel functions in machine learning. 
The purpose of this paper is to clarify roles of Green's functions as a machine learning algorithm. 
In particular, a covariance matrix composed of normalized Green's functions is proposed. 
By supporting Bayesian approach, the covariance matrix provides the mean $\mu$ and the confidence interval 
$[\mu-2s, \mu+2s]$, 
where $s$ stands for a standard deviation.

\section{Green's function}

We start with the following boundary value problem of 2nd order linear ordinary differential equation:
\begin{align}\label{eq:BVP}
\begin{cases}
-\,\dfrac{d^2u}{dx^2}+a^2 u =f(x) \qquad (0<x<1), \\
u(0)=u(1)=0.
\end{cases}
\end{align}
where $a$ is a nonnegative constant. The solution formula of \eqref{eq:BVP} is given by
\begin{align}\label{eq:uGf}
u(x)=\int_0^1 G(x, y) f(y) dy,
\end{align}
where $G(x, y)$ is a Green's  function defined by 
\begin{align}\label{eq:G}
G(x, y)=\sum^{\infty}_{n=1}\frac{2\sin(n\pi x)\sin(n\pi y)}{(n\pi)^2+a^2}=
\begin{cases} \dfrac{\sinh (a\min(x,y))\sinh(a(1-\max(x,y)))}{a\sinh a} & (a>0) \\
\min(x,y)(1-\max(x,y)) & (a=0).
\end{cases}
\end{align}
Let ${\Bbb H}$ be a function space defined by
\begin{align*}
{\Bbb H}=\{u~|~u,u'\in L^2(0,1),~u(0)=u(1)=0\},
\end{align*}
equipped with an inner product 
\begin{align*}
(u,v)_{\Bbb H}=\int_0^1 (u'(x)v'(x)+a^2u(x)v(x))dx.
\end{align*} 
It should be noted that $({\Bbb H}, (\cdot,\cdot)_{\Bbb H})$ is a Hilbert space. 
Kametaka {\it et al.} showed that $G(x,y)$ is a reproducing kernel of 
${\Bbb H}$.\cite{kame,kame2} In other words, the following two properties hold:
\begin{enumerate}
\item[(i)] If one fixes $y \in [0,1]$, $G(x,y)$, as a function of $x$, belongs 
to ${\Bbb H}$.
\item[(ii)] For all $u\in {\Bbb H}$, the following reproducing relation holds:
\begin{align}
(u,G(\cdot,y))_{\Bbb H}=\int_0^1 (u'(x)\partial_xG(x,y)+a^2u(x)G(x,y))dx=u(y).
\end{align}
\end{enumerate}
We consider the case  $a>0$. Since $G(x, y)$ is nonnegative, 
$L^{1}$-norm of a cross section Green function $G(x,y)$ is calculated as follows:
\begin{align*}
L_{1}&=L_{1}(y)=\int_{0}^{1} |G(x, y)|dx = \int_{0}^{1} G(x, y)dx \\
&=\frac1{a\sinh a}\left\{
\int_{0}^{y} \sinh(a x) \sinh(a(1-y))dx+\int_{y}^{1} \sinh(ay)\sinh(a(1-x))dx\right\}\\
&=\frac1{a^{2}}\left(1-\cosh(ay)+\frac{(\cosh a -1)\sinh (ay)}{\sinh a}\right).
\end{align*}
In particular, if $y=\frac12$, we have
\begin{align*}
L_{1}\left(\frac12\right)=\frac{1}{a^{2}}\left(1-\frac1{\cosh\frac{a}{2}} \right)
\end{align*}
We also define $H(x,y)$ as Green's function divided by its $L^1$norm
\begin{align}
H(x,y)=\frac{G(x, y)}{L_{1}(y)} \qquad (0<y<1),
\end{align}
which satisfies the relation
\begin{align}\label{kikaku}
\int_{0}^{1} H(x,y)dx=1.
\end{align}

\begin{figure}[htbp]
\includegraphics{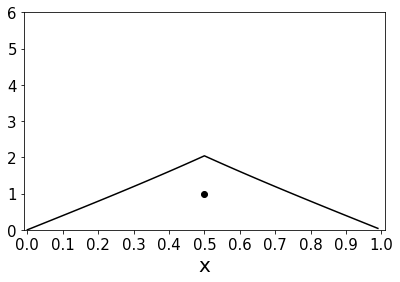}
\caption{Example of  the normalized Green's function, $H(x, 0.5)$ of $a=1$, whose mean value 
$\mu={\Bbb E}[{\bf x}]=0.5$ and standard deviation $s=0.203$. 
The plobability $p(0.5-s \leq x \leq 0.5 +s)=0.652$ and $p(0.5-2s \leq x \leq 0.5 + 2s)=0.965$.}
\label{f1}
\end{figure}

We call the function $H(x,y)$ the normalized Green's function hereafter.
$H(x, 0.5)$ is shown in Fig. \ref{f1}, 
which means the response function by the impulse at point $x=0.5$ and is symmentric in this case.
Note that $H(0, y)=H(1, y)=0$ in accordance with the boundary conditon. 
Here we assume that $H(x,y)$, as a function in $x$, plays a role of a probability density 
function for impulse response. 
By using the function $H(x,y)$, it is expected that one can obtain covarince matrix, 
together with the distribution, mean, and variance. 

Assume that we are given $N$ data sets as follows:
\begin{align}
D=&{ (\xi_1, \eta_1), (\xi_2, \eta_2), \cdots ,(\xi_N, \eta_N)} \label{eq:D}\\
{\boldsymbol \xi}&=(\xi_1, \xi_2, \cdots , \xi_N)^T\label{xdata}   \\
{\boldsymbol \eta}&=(\eta_1, \eta_2, \cdots , \eta_N)^T=(f(\xi_1), f(\xi_2), \cdots, f(\xi_N))^T \label{ydata}
\end{align}
By discretizing the solution formula \eqref{eq:uGf}, the solution $u(x)$ is approximated as 
\begin{align}
u(x)=\sum^{N}_{i=1}G(x, \xi_i) f(\xi_i)\delta,
\label{uG}
\end{align}
where $\delta$ is interval length with respect to $x$.

The relation between machine learning and reproducing kernel theory is pointed out.\cite{fuku} 
Moreover, it is reported the Green's function becomes a reproducing kernel
of a suitable Hilbert space \cite{kame,kame2}. Combining the above results, we can expect 
an application of Green's function theory to machine learning algorithm.
Since the Green's function \eqref{eq:G} is positive, the Green's function divided by its appropriate 
$L^1$ norm plays a role as a probability 
density function and is expected to give a certain regression algorithm. 

From these data sets, we propose a covariance matrix $H$ composed of the normalized Green's functions
as follows:
\begin{align}\label{H}
H=(H_{ij})_{1\leq i, j \leq N}
\end{align}
\begin{align}\label{H2}
H_{ij}&=H(\xi_i, \xi_j)\qquad (i, j=1,2,\cdots , N),
\end{align}
where both of $\xi_i$ and $\xi_j$ are data points.

Many kinds of functions are proposed as entries of $H$~\cite{bishop,gp}.
The essential point of this paper is to adopt the normalized Green's function $H(x, y)$ as the entries 
of covariance matrix $H$. 

Given data $D$, we consider the following problem:\\
Problem~:~Predict $M$ dimensional vector ${\bf y}^{\ast}=(y^{\ast}_1, y^{\ast}_2, \cdots, y^{\ast}_M)^T$ 
at a given point ${\bf x}^{\ast}=(x^{\ast}_1, x^{\ast}_2, \cdots, x^{\ast}_M)^T$. \\ 
We introduce $N+M$ dimensional joint vectors defined by
\begin{align*} 
{\bf x}'=(\xi_1, \xi_2, \cdots, \xi_N, x_1^{\ast}, x_2^{\ast}, \cdots, x_M^{\ast})^T \\
{\bf y}'=(\eta_1, \eta_2, \cdots, \eta_N, y_1^{\ast}, y_2^{\ast},\cdots, y_M^{\ast},)^T,
\end{align*}
where ${\boldsymbol \xi}$ and ${\boldsymbol \eta}$ are data sets $D$ given by Eqs. (\ref{eq:D})-(\ref{ydata}). 
Using Bayesian approach\cite{bishop,gp,kana}, 
$p({\bf y}^{\ast}|{\bf x}^{\ast}, D)$, the predictive distribution  of ${\bf y}^{\ast}$, is 
evaluated from framework of conditional distribution $p({\bf y}^{\ast}|{\boldsymbol \eta})$ based on 
covariance matirx of joint distribution $p({\bf y}')$.  

We here define the $(N+M) \times (N+M)$ covariance matrix $H'$. 
By using the Bayesian approach,\cite{bishop,gp, kana} 
we propose the predictive distribution $p({\bf y}^{\ast}| {\bf x}^{\ast}, D)$ of ${\bf y}^{\ast}$ 
given by fixing ${\boldsymbol \xi}$ and ${\boldsymbol \eta}$ to the observed value of data sets $D$. 
The corresponding covariance matrix $H'$ is given as follows:
\begin{align}
H'=
\begin{pmatrix}\label{HH}
~H~ &~ h_{\ast}~ \\
~ h_{\ast}^T~ & ~ h_{\ast \ast}
\end{pmatrix},
\end{align}
where $H$ is $N \times N$ matrix, $h_{\ast} $ is $N\times M$ matrix and $h_{\ast \ast}$ is $M\times M$ matrix.
The $(i, j)$-th entry of  $  h_{\ast} $  and  $(i, i')$-th entry of  $  h_{\ast \ast } $ are given as 
\begin{align}
h_{\ast}(i, j)=H(x_i, x^{\ast}_j) \quad (i=1,~2, \cdots, N, ~~j=1,2,\cdots, M)\\
h_{\ast \ast}(i, i')=H(x^{\ast}_i, x^{\ast}_{i'}) \quad (i, i'=1,~2, \cdots, M).
\end{align}
By using the above matrices, we find that $M$ dimensional mean vector ${\boldsymbol \mu}$ 
and $M\times M$ covariance matrix $\Sigma$ of 
the predictive distribution $p(\bf y^{*}| {\bf x}^{\ast}, D)$ are given as follows:
\begin{align}
{\boldsymbol \mu}={\Bbb E}[{\bf y}^{\ast}|{\bf x}^{\ast}, D] = h_{\ast}^TH^{-1}{\bf y},
\label{ymu2}
\end{align}
\begin{align}
\Sigma={\rm cov} [{\bf y}^{\ast} | {\bf x}^{\ast}, D] = h_{\ast \ast} - h_{\ast }^TH^{-1}h_{\ast}. 
\label{ysig}
\end{align}
The diagonal entries of covariance matrix $\Sigma$ is equivalent to the predictive variance vector 
${\bf V}=(V_1, V_2, \cdots, V_M)^T$ 
with its $i$-th entry given by:
\begin{align}
V_i= H(x_i^{\ast}, x_i^{\ast}) - {\bf h}_{i}^TH^{-1}{\bf h}_{i} \qquad (i=1, 2, \cdots, M), 
\label{yv2}
\end{align}
where ${\bf h}_i$ is a $i$-th column vector of the matrix $h_{\ast}$.
Note that each entry of the mean vector ${\boldsymbol \mu}$  
and variance vector ${\bf V}$ 
is point-wise function of $x^{\ast}_i$. 
We also introduce standard deviation vector 
${\bf s}=(s_{1},\cdots,s_{M})^{T}$ defined by $s_i=\sqrt{V_i} \quad (i=1,2,\cdots , M)$.

\section{Results}
In this section, we present numerical results concerning the application of Green's function to a regression algorithm.
We consider two cases $a=1$ and $a=10$ in the differential equation (\ref{eq:BVP}).
We also put the difference interval of $[0,1]$ as $\delta = 0.01$, or equivalently $M=100$, 
throughout this section. 
We also give $N=5$ data sets as follows:
\begin{align*}
D=(0.1, 1), (0.3, 2), (0.5, 3), (0.7, 4), (0.9, 5),  \\
{\boldsymbol \xi}=(0.1, 0.3, 0.5, 0.7, 0.9),\quad {\boldsymbol \eta}=(1, 2, 3, 4, 5).
\end{align*}

\begin{figure}
\begin{center}
\includegraphics{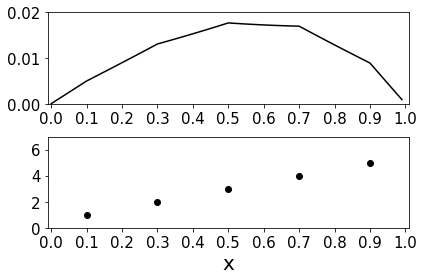}
\end{center}
\caption{The solution $u(x)$ of differential equation in  the case of $a=1$. 
The soild line in upper part is $u(x)$ calculated by Eq. (\ref{uG}). Solid circles in  lower part represent data sets $D$.}
\label{f2}
\end{figure}

\begin{figure}
\includegraphics{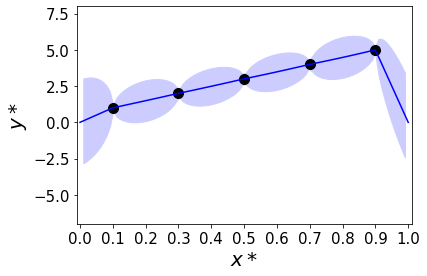}
\caption{The predictive distribution $p({\bf y}^{\ast}| {\bf x}^{\ast}, D)$ of $y^{\ast}$ at $x^{\ast}$ 
generated by 5 data points $D$ in the case of $a=1$.  
The solid line shows the predictive mean $\mu$. 
The shaded region corresponds to confidence interval with its width plus and minus $2s$. 
Solid circles are data sets $D$.}
\label{f3}
\end{figure}

We first consider the case $a=1$. 
Figure \ref{f1} shows $H(x, 0.5)$, which is a cross section of normalized
Green's function. 
This looks like a straight line in this case. 
Upper part of Fig. \ref{f2}
shows the discretized solution $u(x)$ produced by linear combination, 
\begin{align*}
u(x)=\displaystyle \sum_{j=1}^{5} \eta_j G(x, x_j)\delta.
\end{align*}
Lower part shows original 5 data points.  
The solid line of $u(x)$ reflects the original data points with $u(0)=u(1)=0$.
In Fig. \ref{f3}, 
the solid line corresponds to the predictive mean $\mu$ 
as a point-wise function of $x^{\ast}$. 
The shaded region spans from $\mu-2s$ to $\mu + 2s$ in the vertical direction, 
where the standard deviation $s=\sqrt{V}$ is 
given from the predictive variance $ V$ as a function of $x^{\ast}$, 
and corresponds to confidence interval.\cite{chen}
It is observed that the span of the shaded region depends on $x^{\ast}$ 
and is the smallest in the neighbourhood of the data points. 
Substituting data sets $D$ to ${\boldsymbol \xi}$ and ${\boldsymbol \eta}$, 
we obtain a covariance matrix $H'$, composed of normalized Green's function $H(x,y)$, in Eq. (\ref{HH}). 
The block matrix $H$ in $H'$ is given by 
\begin{align}\label{Ha1}
H(a=1)=
\begin{pmatrix}
2.119 & 0.678 & 0.392 & 0.272 & 0.207 \\
1.566 & 2.061 &  1.193 & 0.827 &  0.629 \\
1.076 & 1.416 &  2.041 & 1.416 & 1.076 \\
0.629 & 0.827 &  1.193 & 2.061 & 1.566 \\
0.207 & 0.272 &  0.392 & 0.678 & 2.119
\end{pmatrix}.
\end{align}
By Bayesian approach the covariance matrix gives the predictive distribution $p({\bf y}^{\ast}| {\bf x}^{\ast}, D)$ of ${\bf y}^{\ast}$, 
from which one can find its predictive mean $\mu$ and variance $V$ as is shown in Fig. \ref{f3}.

\begin{figure}
\includegraphics{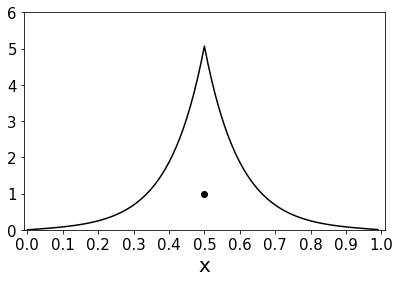}
\caption{Example of  the normalized Green's function, $H(x, 0.5)$ of $a=10$, 
whose mean value $\mu={\Bbb E}[{\bf x}]=0.5$ and standard deviation $s=0.129$. 
The plobability $p(0.5-s \leq x \leq 0.5 +s)=0.734$ and $p(0.5-2s \leq x \leq 0.5 + 2s)=0.936$.}
\label{f4}
\end{figure}

\begin{figure}
\includegraphics{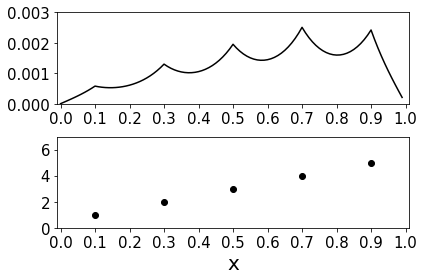}
\caption{The solution $u(x)$ of differential equation in  the case of $a=10$. 
The soild line in upper part is $u(x)$ calculated by Eq. (\ref{uG}). Solid circles in lower part represent data sets $D$.}
\label{f5}
\end{figure}

We next consider the case $a=10$. 
Figure \ref{f4} shows $H(x, 0.5)$, which has a curved line, a pointed peak, and a 
narrower distribution compared with that in the case of $a=1$.
Figure \ref{f5} shows the discretized solution $u(x)$. 
The solid line of $u(x)$ also reflects the original data points, 
however the tendency of pointed peak looks around data points.
Figure \ref{f6} illustrates the predictive destribution $p({\bf y}^{\ast}| {\bf x}^{\ast}, D)$. 
The solid line represents the mean reflecting the pointed peak. 
Note that the shaded region,
corresponding to confidence interval\cite{chen} of plus and minus $2s$, stretches wider 
than that in the case of $a=1$ in Fig. \ref{f3}.
The block matrix $H$ in $H'$ is given by 
\begin{align}\label{Ha10}
H(a=10)=
\begin{pmatrix}
6.841 & 0.616 & 0.080  & 0.011 &  0.002 \\
0.926 & 5.254 &  0.684 &  0.096 &  0.017 \\
0.125 & 0.711 &  5.068 & 0.711 &  0.125 \\
0.017 & 0.096 &  0.684 &  5.254 &  0.926 \\
0.002 & 0.011 &  0.080 &  0.616 & 6.841
\end{pmatrix}.
\end{align}

\section{Discussions}

First we compare normalized Green's functions $H(x,y)$ which is regarded as plobability density functions. 
Examples of $H(x,0.5)$ in the case of $a=1$ and $a=10$ are shown in Figs. \ref{f1} and \ref{f4}, and in Table \ref{T1}. 
In both cases, $H(x,0.5)$  are symmetric, therefore the means of $x$ are 0.5. 
Table \ref{T1} shows that the variances of $x$ are different in the case $a=1$ and $a=10$.
In the case of $a=1$, the variance ${\rm var}[X]=0.041$, and standard deviation $s= 0.203$,  
so probability $p(0.5- s\leq x \leq 0.5+s)=0.652$ and $p(0.5-2s\leq x \leq 0.5+2s)=0.965$. 
In the case of $a=10$, ${\rm var}[X]=0.017$, $s= 0.129$,  
probability $p(0.5- s\leq x \leq 0.5+s)=0.734$, and $p(0.5-2s\leq x \leq 0.5+2s)=0.936$. 
It is observed that the length of standard deviation $s$ in  the case of $a=1$ is wider than that of $a=10$. 
The probability $p(0.5-s\leq x \leq 0.5+s)$ of $a=1$ is lower than that of $a=10$, 
however the magnitude relation of $p(0.5-2s\leq x \leq 0.5+2s)$ is reversed. 
This means that the correlation by Green's function concentrates in the narrower range as $a$ gets larger. 

\begin{figure}
\includegraphics{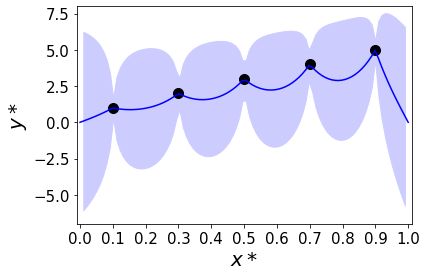}
\caption{The predictive distribution $p({\bf y}^{\ast}| {\bf x}^{\ast}, D)$ of $y^{\ast}$ at $x^{\ast}$ 
generated by 5 data points $D$ in the case of $a=10$. 
The solid line shows the predictive mean $\mu$. 
The shaded region corresponds to confidence interval with its width plus and minus $2s$. 
Solid circles are data sets $D$.}
\label{f6}
\end{figure}

Next we compare the span of shaded region in the vertical direction 
as a point-wise function of $x^{\ast}$ of Figs. \ref{f3} and \ref{f6}, 
which corresponds to confidence interval.\cite{chen} 
Here we check $5\times5$ block matrix $H(a=1)$ of Eq. (\ref{Ha1}) and $H(a=10)$ of Eq. (\ref{Ha10}), 
which are the covariance matrix for ${\boldsymbol \xi }=(0.1,~0.3,~0.5,~0.7,~0.9)$ of data sets $D$. 
In Table \ref{T1}, we compare the value of $H(\xi_3, \xi_3)$ and  $H(\xi_3, \xi_1)$ for example. 
The diagonal value $H(\xi_i, \xi_i)$ gets larger and $H(\xi_i, \xi_j)$ $(\xi_i \neq \xi_j)$ gets lower as $a$ gets larger. 
Due to the contribution of diagonal term $H(x^{\ast}_{i}, x^{\ast}_{i})$, 
the first term of Eq. (\ref{yv2}) is more dominant as $a$ gets larger. 
For example, numerical calculation at $x^{\ast}=0.4$ shows that 
first term $H(0.4,0.4)$ is 2.046 and second term ${\bf h}_{i}^{T}H^{-1}{\bf h}_{i}$ is 1.661 in the case 
of $a=1$, whereas first term is 5.101 and second term is 1.233 in the case of $a=10$. 
Therefore the span of shaded region in the vertical direction at $x^{\ast}$ of $a=1$ in Fig. \ref{f3} is narrower than that of $a=10$ in Fig. \ref{f6}. 
In other words, wider standard deviation 
causes the narrower span of shaded region, which corresponds to narrower confidence interval. 

\begin{table}
\caption{Values of the noremalized Green's function of $a=1$ and $a=10$.}
\label{T1}
\begin{center}
\begin{tabular}{lll}
\hline
\multicolumn{1}{c}{values } & \multicolumn{1}{c}{$a=1$} & \multicolumn{1}{c}{$a=10$} \\
\hline
mean$:\mu={\Bbb E}[X]$ &  0.5 & 0.5 \\
variance: $V={\rm var}[X]$ & 0.041 & 0.017  \\
standard deviation:$s=\sqrt{V}$ & 0.203  &  0.129  \\
plobability:$p(0.5-s\leq x \leq 0.5+s)$ &0.652 & 0.734   \\
plobability:$p(0.5-2s\leq x \leq 0.5+2s)$ & 0.965 & 0.936 \\
value:$H(\xi_3, \xi_3)$  & 2.041  & 5.068  \\
value:$H(\xi_3, \xi_1)$  & 1.076 &  0.125 \\
\hline
\end{tabular}
\end{center}
\end{table}

\section{Concluding remarks}

In this paper we proposed and implemented a regression algorithm based on Green's function theory. 
The estimation of $u(x)$ by using Green's function reflects data points, 
having a reproducing kernel of a suitable Hilbert space. 
This implies the possibility of Green's functions as probability density functions. 
Regarding a normalized Green's function as the probability density function, 
we considered a covariance matrix. 
By Bayesian approach, the covariance matrix gives a predictive distribution, 
which produces mean and variance.

\section*{Acknowledgment}
The author would like to thank S. Kamei and I. Kayo of Tokyo University of Technology, S. Tomizawa of Toyota Technological Institute, and T. Miura of National Institute of Advanced Industrial Science and Technology for useful comments.


\end{document}